\documentclass{IOS-Book-Article}

\usepackage{mathptmx}
\usepackage{soul}\setuldepth{article}
\usepackage{booktabs}
\usepackage{multirow}
\usepackage{tcolorbox}
\usepackage{soul}
\newcommand{\ctext}[3][RGB]{%
  \begingroup
  \definecolor{hlcolor}{#1}{#2}\sethlcolor{hlcolor}%
  \hl{#3}%
  \endgroup
}

\newtcolorbox{hs}[3][]
{
  colframe = black,
  colback  = white,
  coltitle = #2!20!black,  
  sharp corners = southwest,
  arc=2mm,
  boxrule=1pt,
  left=2pt,
  right=2pt,
  top=2pt,
  bottom = 2pt,
  middle = 2pt,
  #1,
}
\def\hb{\hbox to 11.5 cm{}}

\begin{document}

\pagestyle{headings}
\def\thepage{}
\begin{frontmatter}              

\title{Leveraging Argument Structure to Predict Content Hatefulness}

\markboth{}{April 2026\hb}

\author[A]{\fnms{Nicolás Benjamín} \snm{Ocampo}\orcid{0009-0001-0077-4626}%
\thanks{Corresponding Author: Nicolás Benjamín Ocampo, Science Park, 123, 1098XG, Amsterdam, The Netherlands}} and
\author[A]{\fnms{Davide} \snm{Ceolin}\orcid{0000-0002-3357-9130}}

\runningauthor{B.P. Manager et al.}
\address[A]{Centrum Wiskunde \& Informatica, Amsterdam, The Netherlands}

\begin{abstract}
Information disorder is a challenging phenomenon that affects society at large. This phenomenon entails the diffusion of misleading, misinforming, and hateful content online. In different contexts, one aspect of the problem may prevail, but overall, this is a broad problem that requires comprehensive solutions. While each dimension of the problem (hate speech, disinformation, misinformation, etc.) requires in-depth analysis, in this paper, we look into the possibility of argument structure to provide relevant information to link these different areas of the problem. In particular, we focus on the WSF-ARG+ dataset, which consists of white supremacy forum messages annotated in terms of argument structure (premises and conclusion). There, we leverage the checkworthiness and hatefulness annotations of the argument components to obtain insights into the hatefulness of the whole message.
Our results show promising insights (up to 96\% F1), indicating the possibility of extending this direction in the future to tackle hateful content identification and information disorder countering.
\end{abstract}

\end{frontmatter}
\markboth{April 2026\hb}{April 2026\hb}

\noindent {\color{red} \textbf{Content Warning}: This paper contains examples of language which may be offensive.}

\section{Introduction}

The information disorder~\cite{WardleDerakhshan2017} is a well-recognized phenomenon that encompasses misleading and harmful content being generated and shared online, intentionally or not.
Misinformation and hate speech (HS) represent two main dimensions in the information disorder; they represent significant challenges and are therefore usually tackled separately in the literature. Nevertheless, misinformation and hatefulness often concur and reinforce one another \cite{biradar-etal-2024-proceedings}. Hateful content can be reinforced by fact-like (but tendentious) content, while misinformation can leverage a charged tone and hateful content to spread more quickly. For the same reason, much of this hateful and misinforming content follows an argumentative structure, because this gives malicious content a more solid appearance.
This paper focuses on the intersection of these three domains (argumentation, misinformation countering, and hate speech detection). In particular, we investigate whether the argumentative structure of these messages can help in predicting their hatefulness, knowing that the argumentative components (premises and conclusions) are characterized in terms of checkworthiness and hatefulness. From our experiments on a white supremacy forum dataset, we observe that the information encoded in the premises is sufficient to assess the whole message and that this line of research show promising results.
The rest of the paper is structured as follows. Section~\ref{sec:related} presents related work. Our method is presented in Section~\ref{sec:methodology} and the results are discussed in Section~\ref{sec:results}. Section~\ref{sec:conclusion} concludes. 

\section{Related Work}
\label{sec:related}
Hate Speech has been represented using multiple datasets covering dimensions such as target groups being attacked, levels of implicitness, hateful taxonomies, among others \cite{kapil-ekbal-2024-survey}. However, very little attention has been given to the their argumentative structure. The ASOHMO corpus \cite{furman-etal-2023-argumentative} was the first attempt to identify argumentative aspects in hateful messages from Twitter, annotating their \textit{justification} and \textit{conclusion}. The ASOHMO corpus was later used by \textit{Furman et al.} \cite{furman-etal-2023-high} to generate counterspeech, showing that focusing on argument components yields higher-quality results than using the full message. 
The main limitation of this dataset is that premises and conclusion might not be stand-alone sentences; justifications and conclusion can consist of only hashtags (e.g. \textit{\#buildthedamnwall}). 
WSF-ARG \cite{bonaldi-etal-2024-safer} addresses this with messages long enough to likely contain an argumentative structure releasing 227 argumentative hateful messages from the white supremacy forum Stormfront, later used to generate more cogent and less repetitive counterspeech. 
Similarly, Saha et al. \cite{saha-srihari-2024-integrating} extract argumentative graphs from counterspeech dialogues to fine-tune a model for generating counterspeech on MisinfoCorrect \cite{he2023reinforcement}, a COVID-19 vaccine misinformation dataset.

At the intersection of hate speech and misinformation, WSF-ARG+ \cite{ocampo2026hatemeetsfactsllmsintheloop} extends WSF-ARG by adding argumentative non-hate speech messages and check-worthiness annotations on the argument components of both hateful and non-hateful messages. Check-worthiness, referring to verifiable assertions of public interest or impact relevant for fact-checking, is found to apply to 49.60\% of the argument components in hateful messages in WSF-ARG+. Faux-hate \cite{biradar-etal-2024-proceedings} extracts fake narratives from reliable fact-checkers to use them as queries to retrieve reactions from Twitter and YouTube. These reactions are then annotated with hate/non-hate, fake/non-fake, and severity labels. 


\section{Methodology}
\label{sec:methodology}

An argument can be modeled as a set of premises followed by a conclusion they are intended to support. The overall hatefulness of a message depends in part on the hatefulness of its individual components, but it is not simply the sum of these elements. It is also shaped by how reasoning links premises to the conclusion. To predict a message’s hatefulness, we assess each argument component individually and study how different sequences of hateful and non-hateful premises and conclusions affect the whole. We also consider the check-worthiness of each argument component, which does not determine their veracity, but it provides insight into the type of reasoning, helping distinguish between verifiable and subjective premises and conclusions.
E.g., the following message is composed of \ctext[RGB]{255,187,158}{premise 1}, \ctext[RGB]{181,228,224}{premise 2}, and \ctext[RGB]{255,234,148}{conclusion}:
\begin{hs}{black}{}{}
It is important for people to realize that \ctext[RGB]{255,187,158}{gays want gay marriage} ..... \ctext[RGB]{181,228,224}{not because they desire some sort of government paperwork}, but because \ctext[RGB]{255,234,148}{they are seeking the right to adopt defenseless children}.
\end{hs}
Both premises 1 and 2 are, on their own, non-hateful and checkworthy. However, they support and reinforce a conclusion that is both check-worthy and hateful.


\subsection{Data}

\begin{table}[b]
\centering
\resizebox{!}{2.132cm}{\
\begin{tabular}{l|cccc|cc|cc}
\toprule
& \multicolumn{4}{c|}{Hateful Messages} & \multicolumn{2}{c|}{Non-Hateful Messages} &  &                      \\ \cline{2-9}
& \multicolumn{2}{c}{Premises} & \multicolumn{2}{c|}{Conclusion} & & & & \\
\multirow{-2}{*}{\shortstack[l]{Argument \\ Components}} & Non-HS & HS & Non-HS & HS & \multirow{-2}{*}{Premises} & \multirow{-2}{*}{Conclusion} & \multicolumn{2}{c}{\multirow{-2}{*}{ALL}} \\ \hline
NFS & 29  & 45  & 30 & 98 & 107 & 105 & 414 & \\
UFS & 70  & 29  & 7  & 11 & 160 & 13 & 290 & \\
CFS & 110 & 123 & 21 & 60 & 94  & 18 & 426 & \multirow{-3}{*}{1130}\\ \midrule
    & 209 & 197 & 58 & 169 & 361 & 136 & & \\ \cline{2-7}
    & \multicolumn{2}{c}{406} & \multicolumn{2}{c|}{227} &  &  & &  \\ \cline{2-5}
\multirow{-3}{*}{ALL} & \multicolumn{4}{c|}{633} & \multicolumn{2}{c|}{\multirow{-2}{*}{497}} & & \\ 
\bottomrule
\end{tabular}}
\vspace{0.5em}
\caption{Distribution across argument components, hatefulness, and check-worthiness labels in WSF-ARG+.}
\label{tab:label-dist-wsf-arg-plus}
\end{table}

In order to focus on the intersection of argumentation, misinformation countering, and hate speech detection, we relied on the WSF-ARG+ dataset consisting of 227 hateful and 136 non-hateful argumentative messages, i.e., containing a conclusion supported by one or more premises (with an average of 1.789 $\pm$ 0.644 premises for hateful messages and 2.654 $\pm$ 1.157 for non-hateful ones). The dataset accounts for several different annotation features, such as their extracted argument components, whether the argument components are checkworthy, and for the hateful messages, which of their components can be considered hateful when looked at in isolation.




For the checkworthiness annotation layer, the dataset follows the ClaimBuster framework \cite{claimbuster}, which defines three labels: Checkworthy Factual Statement (CFS), Unimportant Factual Statement (UFS), and Non-Factual Statement (NFS). These labels indicate whether a claim is important and should be fact-checked, whether it is factual but of low interest to the general public, or whether it is subjective or opinion-based, respectively. Table \ref{tab:label-dist-wsf-arg-plus} summarizes the label distribution of WSF-ARG+ intersecting hatefulness, checkworthiness, and argument component annotations.

\subsection{Approach}

We characterize our problem as a hate speech classification task, and test a range of classifiers in their ability to predict the hatefulness of a message based on the hatefulness and checkworthiness of its argumentative components. A message is encoded as:


\paragraph{Argumentative Structure (arg-str)} 
Each message is represented as an ordered sequence of its components from left to right, with the premises occupying the first positions and the conclusion the final one. We encode this structure using a fixed-length one-hot vector, 
where each position corresponds to a potential component slot. A value of \textit{1} indicates the presence of a component in that position, and \textit{0} otherwise. Since all messages are argumentative, each instance contains at least one premise and one conclusion, resulting in at least two active positions.
  
\paragraph{Premise-only Argumentative Structure (agr-str-p)} It follows the same encoding as \textit{arg-str} but excludes the conclusion component. The resulting one-hot vector captures only the argumentative structure of the premises quantifying the extent to which premise-level information alone contributes to the detection task.
 
\paragraph{Conclusion conditioned on Premises (c-given-p)} 
A model is first trained using only the one-hot encoding of the premises to generate hatefulness predictions. These predictions are then concatenated with the one-hot encoding of the conclusion. This combined representation lets us evaluate the conclusion’s contribution given the premise predictions.

\paragraph{Checkworthiness Annotations (cw)} To jointly encode structural and checkworthiness features, each of the previous representations is extended with a one-hot vector indicating the argument component labels (NFS, UFS, CFS). The ordering retains the original left-to-right structure, with premise components placed before the conclusion. 

\paragraph{Hateful Annotations (hs)} To incorporate hate speech annotations into the argument components, we extend both the structural encodings and the checkworthiness encodings by adding a binary vector for each component, where 1 indicates that the component is hateful and 0 indicates that it is non-hateful. Missing annotations are encoded as 0.

To study the predictive performance of these encodings, we employ a set of small classifiers: \textit{Logistic Regression (lgr)}, \textit{Support Vector Machines (svm)}, \textit{Random Forest (rforest)}, and \textit{XGBoost (xgb)}. We train each model on binary hate speech detection (HS vs Non-HS) using WSF-ARG+ messages (227 HS, 136 Non-HS). We use a 5-Fold stratified cross validation to test across several splittings of the dataset. Models use \texttt{max\_iter=1000}, \texttt{random\_state=0}, and log loss as the optimization objective.
 

%

\begin{table}[t]
\centering
\begin{tabular}{llccc}
\toprule
Encoding & Model & Precision & Recall & Macro F1 \\
\midrule
\multirow[t]{4}{*}{arg-str} & lgr & 0.730 ± 0.066 & 0.696 ± 0.068 & 0.701 ± 0.074 \\
 & rforest & 0.730 ± 0.066 & 0.696 ± 0.068 & 0.701 ± 0.074 \\
 & svm & 0.726 ± 0.064 & 0.693 ± 0.066 & 0.698 ± 0.073 \\
 & xgb & 0.726 ± 0.064 & 0.693 ± 0.066 & 0.698 ± 0.073 \\
\cline{1-5}
\multirow[t]{4}{*}{arg-str-p} & lgr & 0.730 ± 0.066 & 0.696 ± 0.068 & 0.701 ± 0.074 \\
 & rforest & 0.730 ± 0.066 & 0.696 ± 0.068 & 0.701 ± 0.074 \\
 & svm & 0.726 ± 0.064 & 0.693 ± 0.066 & 0.698 ± 0.073 \\
 & xgb & 0.726 ± 0.064 & 0.693 ± 0.066 & 0.698 ± 0.073 \\
\cline{1-5}
\multirow[t]{4}{*}{arg-str-c-given-p} & lgr & 0.730 ± 0.066 & 0.696 ± 0.068 & 0.701 ± 0.074 \\
 & rforest & 0.730 ± 0.066 & 0.696 ± 0.068 & 0.701 ± 0.074 \\
 & svm & 0.726 ± 0.064 & 0.693 ± 0.066 & 0.698 ± 0.073 \\
 & xgb & 0.726 ± 0.064 & 0.693 ± 0.066 & 0.698 ± 0.073 \\
\cline{1-5}
\multirow[t]{4}{*}{arg-str-cw} & lgr & 0.729 ± 0.070 & 0.688 ± 0.062 & 0.693 ± 0.069 \\
 & rforest & 0.667 ± 0.067 & 0.646 ± 0.058 & 0.646 ± 0.063 \\
 & svm & 0.733 ± 0.060 & 0.685 ± 0.058 & 0.690 ± 0.064 \\
 & xgb & 0.660 ± 0.068 & 0.640 ± 0.062 & 0.640 ± 0.066 \\
\cline{1-5}
\multirow[t]{4}{*}{arg-str-p-cw} & lgr & 0.719 ± 0.068 & 0.681 ± 0.065 & 0.685 ± 0.074 \\
 & rforest & 0.672 ± 0.038 & 0.650 ± 0.046 & 0.649 ± 0.053 \\
 & svm & 0.699 ± 0.070 & 0.674 ± 0.065 & 0.674 ± 0.074 \\
 & xgb & 0.672 ± 0.032 & 0.651 ± 0.044 & 0.650 ± 0.051 \\
\cline{1-5}
\multirow[t]{4}{*}{arg-str-c-given-p-cw} & lgr & 0.727 ± 0.076 & 0.681 ± 0.066 & 0.686 ± 0.075 \\
 & rforest & 0.672 ± 0.038 & 0.650 ± 0.046 & 0.649 ± 0.053 \\
 & svm & 0.707 ± 0.076 & 0.666 ± 0.065 & 0.666 ± 0.077 \\
 & xgb & 0.672 ± 0.032 & 0.651 ± 0.044 & 0.650 ± 0.051 \\
\cline{1-5}
\multirow[t]{4}{*}{arg-str-hs} & lgr & 0.940 ± 0.033 & 0.958 ± 0.025 & 0.946 ± 0.032 \\
 & rforest & 0.951 ± 0.024 & 0.967 ± 0.017 & \textbf{0.957 ± 0.023} \\
 & svm & 0.951 ± 0.024 & 0.967 ± 0.017 & \textbf{0.957 ± 0.023} \\
 & xgb & 0.943 ± 0.031 & 0.960 ± 0.023 & 0.949 ± 0.029 \\
\cline{1-5}
\multirow[t]{4}{*}{arg-str-cw-hs} & lgr & 0.931 ± 0.022 & 0.944 ± 0.010 & 0.934 ± 0.018 \\
 & rforest & 0.933 ± 0.026 & 0.944 ± 0.020 & 0.937 ± 0.024 \\
 & svm & 0.938 ± 0.036 & 0.956 ± 0.028 & 0.943 ± 0.036 \\
 & xgb & 0.930 ± 0.030 & 0.941 ± 0.027 & 0.933 ± 0.028 \\
\bottomrule
\end{tabular}
\vspace{0.5em}
\caption{Mean ± standard deviation across 5-fold stratified cross-validation using arg-str, arg-str-p, and c-given-p encodings, their extended encodings with checkworthiness annotations (arg-str-cw, arg-str-p-cw, c-given-p-cw), and with additional hatefulness annotations (arg-str-hs, arg-str-cw-hs).} 
\label{clf:results}
\end{table}

\section{Results \& Discussion}
\label{sec:results}

Table \ref{clf:results} shows the classification results for each encoding strategy and selected model across the five folds. The results indicate that all configurations capture relevant structural information for hate-speech detection, achieving average F1 scores above 0.636, compared to a random baseline (F1 = 0.5). 
Using only the argumentative structure of the messages 
we obtain up to 0.701 F1 on average, while incorporating checkworthiness annotations reduces F1, affecting both precision and recall in all models.
Most of the predictive information is contained in the structure of the premises, obtaining the same detection results on \textit{arg-st}, \textit{arg-str-p}, and \textit{arg-str-c-given-p}. The same effect occurs when checkworthiness is added obtaining comparable results. This is mainly because (i) the dataset includes argumentative messages that contain a single conclusion; therefore, this encoding is always present in both hateful and non-hateful messages, and (ii) only minimal, yet still relevant, features are encoded in the structure of the messages in WSF-ARG+. This does not mean that the structure of the conclusion cannot be relevant for detection, especially in discourses where messages may contain more than one conclusion. Leveraging the argumentative structure together with hate speech annotations on argument components yields the highest performance, achieving an average F1 score of up to 0.957. This configuration also reduces variability across the 5 folds. Although significant benefits can be obtained from hate speech annotations on argument components, ground-truth annotations are not available in practice. However, we believe these results provide an indication of the potential gains achievable by leveraging structural information and encoding cues of hatefulness across different arguments. In contrast, jointly encoding checkworthiness and hatefulness has a detrimental effect on performance. While checkworthiness annotations are relevant and beneficial when combined with both the argumentative structure and textual claims, they are most effective in models with sufficient capacity to properly exploit this additional information, such as large language models (as shown in Ocampo et al. \cite{ocampo2026hatemeetsfactsllmsintheloop}). In smaller models, their inclusion negatively impacts performance on WSF-ARG+.

\section{Conclusion}
\label{sec:conclusion}
In this paper, we focus on the intersection of three domains—argumentation, misinformation, and hate speech detection—where the latter two often co-occur, reinforcing one another within an argumentative structure (i.e., a set of premises leading to a conclusion). Several studies have attempted to address these two dimensions separately, but rarely together. Moreover, these approaches typically rely on the complete message rather than examining its underlying argumentative structure. 
With this in mind, we investigate whether (i) the argumentative structure of messages and (ii) the hatefulness and checkworthiness of each argument component are useful for detecting hate speech in the WSF-ARG+ dataset. We show that, when using the argumentative structure and argument-based hatefulness annotations, we obtain an average F1-score of up to 0.701 and 0.957 in 5-fold cross-validation, respectively.
Moreover, while checkworthiness annotation can significantly improve hate speech detection with large language models, as shown by \textit{Ocampo et al.} \cite{ocampo2026hatemeetsfactsllmsintheloop}, they can have a detrimental effect on smaller models, needing to be used with both the argumentative structure, the claims, and models with large enough contexts. We believe this work sets a starting point for analyzing the argumentative structures and verifiable claims present in hate speech datasets.

\bibliographystyle{vancouver}
\bibliography{custom}



\end{document}